\documentclass[sigconf]{acmart}

\usepackage{multirow}
\usepackage{booktabs}
\usepackage{algorithm}
\usepackage{algorithmic}
\usepackage{graphicx}
\usepackage{arydshln}
\usepackage{amsmath}
\usepackage{amsthm}

\usepackage{subfloat}
\usepackage{subfig}

\usepackage{fancyhdr}
\AtBeginDocument{%
	}

\AtBeginDocument{%
	}

\setcopyright{acmcopyright}
\copyrightyear{2018}
\acmYear{2018}
\acmDOI{XXXXXXX.XXXXXXX}

\acmConference[MM '24]{Proceedings of the 31th ACM International Conference on Multimedia}{October 28--November 01,
	2024}{Melbourne, Australia}

\acmPrice{15.00}
\acmISBN{978-1-4503-XXXX-X/18/06}

\renewcommand\footnotetextcopyrightpermission[1]{}
\begin{document}
	
\title{Revisiting Multimodal Emotion Recognition in Conversation from the Perspective of Graph Spectrum}

\author{Tao Meng}
\affiliation{
	\institution{Central South University of Forestry and Technology}
	\city{Hunan}
	\country{China}}
\email{mengtao@hnu.edu.cn}

\author{Fuchen Zhang}
\affiliation{
	\institution{Central South University of Forestry and Technology}
	\city{Hunan}
	\country{China}}
\email{fuchen.zhang@csuft.edu.cn}

	\author{Yuntao Shou}
	\authornote{Corresponding Author}
\affiliation{
	\institution{Central South University of Forestry and Technology}
	\city{Hunan}
	\country{China}}
\email{shouyuntao@stu.xjtu.edu.cn}

\author{Wei Ai}
\affiliation{
	\institution{Central South University of Forestry and Technology}
	\city{Hunan}
	\country{China}}
\email{aiwei@hnu.edu.cn}

\author{Nan Yin}
\affiliation{
	\institution{Mohamed bin Zayed University of Artificial Intelligence
	}
	\country{UAE}}
\email{nan.yin@mbzuai.ac.ae}

\author{Keqin Li}
\affiliation{
	\institution{State University of New York, New Paltz, New York 12561, USA}
	\city{Xi'an}
	\country{China}}
\email{lik@newpaltz.edu}

\renewcommand{\shortauthors}{Trovato et al.}
	
\begin{abstract}
Efficiently capturing consistent and complementary semantic features in a multimodal conversation context is crucial for Multimodal Emotion Recognition in Conversation (MERC). Existing methods mainly use graph structures to model dialogue context semantic dependencies and employ Graph Neural Networks (GNN) to capture multimodal semantic features for emotion recognition. However, these methods are limited by some inherent characteristics of GNN, such as over-smoothing and low-pass filtering, resulting in the inability to learn long-distance consistency information and complementary information efficiently. Since consistency and complementarity information correspond to low-frequency and high-frequency information, respectively, this paper revisits the problem of multimodal emotion recognition in conversation from the perspective of the graph spectrum. Specifically, we propose a Graph-Spectrum-based Multimodal Consistency and Complementary collaborative learning framework GS-MCC. First, GS-MCC uses a sliding window to construct a multimodal interaction graph to model conversational relationships and uses efficient Fourier graph operators to extract long-distance high-frequency and low-frequency information, respectively. Then, GS-MCC uses contrastive learning to construct self-supervised signals that reflect complementarity and consistent semantic collaboration with high and low-frequency signals, thereby improving the ability of high and low-frequency information to reflect real emotions. Finally, GS-MCC inputs the collaborative high and low-frequency information into the MLP network and softmax function for emotion prediction. Extensive experiments have proven the superiority of the GS-MCC architecture proposed in this paper on two benchmark data sets.
\end{abstract}



\maketitle

\section{Introduction}
\label{sec:introduction}
\begin{figure}[htbp]
	\centering
	\includegraphics[width=1.0\linewidth,scale=1.00]{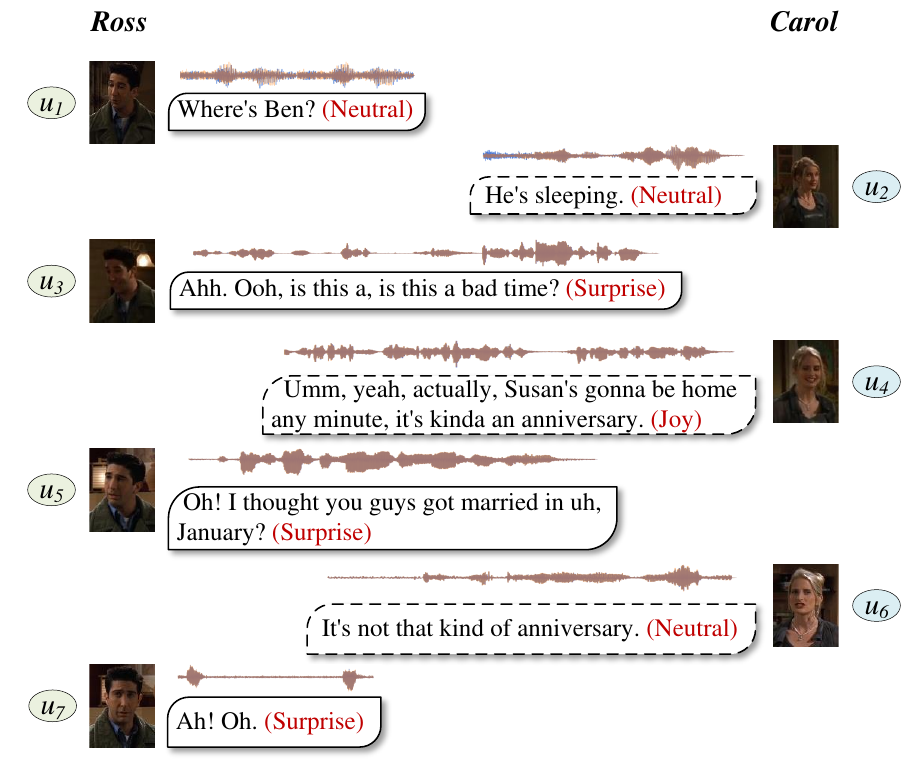}
	\caption{An example of a multimodal conversation from the MELD dataset. MERC aims to identify each utterance's emotion label (e.g., \textit{Neutral, Surprise, Joy}).}
	\label{fig_1}
\end{figure}

With the continuous development of Human-Computer Interaction (HCI), the multimodal emotion recognition task in conversation (MERC) has recently received extensive research attention \cite{majumder2019dialoguernn, ghosal2019dialoguegcn, ai2024gcn, cheng2023semi, liu2023emotionkd, wu2022leveraging, yang2022disentangled}. MERC aims to identify the emotional state of each utterance using textual, acoustic, and visual information in the conversational context \cite{lian2023gcnet, yang2024emotion, shou2022conversational, shou2023comprehensive, meng2023deep}, which is crucial for multimodal conversational understanding and an essential component for building intelligent HCI systems \cite{hu2021mmgcn, mai2022hybrid, meng2024multi}. As shown in Fig. \ref{fig_1}, MERC needs to recognize the emotion of each multimodal utterance in the conversation.

Unlike traditional unimodal or non-conversational emotion recognition \cite{gerczuk2021emonet, deng2021survey, shou2023adversarial, ai2024gcn}, MERC requires joint conversational context and multimodal information modeling to achieve consistency and complementary semantic capture within and between modalities \cite{zhang2024multi}. Fig. \ref{fig_1} gives an example of a multimodal conversation between two people, Ross and Carol, from the MELD dataset. As shown in utterance $u_4$, Carol has a ``Joy'' emotion, which is vaguely reflected in textual features but more evident in visual or auditory features reflecting the complementary semantics between modalities. In addition, it is difficult to identify the emotion of ``Surprise'' from the utterance $u_7$ alone. However, due to the potential consistency of conversational emotions, it can be accurately inferred based on previous utterances. Therefore, the key to multimodal conversational emotion recognition is to capture the consistency and complementary semantics between multimodal information by utilizing the conversational context and emotional dependence between speakers to reveal the speaker's genuine emotion.

The current mainstream research method uses the Transformer \cite{lian2021ctnet, ma2023transformer, zou2023multimodal, zhao2023tdfnet} or GNN \cite{li2023graphcfc, li2023revisiting, tu2024adaptive, ai2023two} architecture to model the MERC task. Transformer-based methods mainly learn complex semantic information between multimodal and conversational contexts from global sequence modeling. For example, CTNet \cite{lian2021ctnet} builds a single Transformer and cross Transformer to capture long-distance context dependencies and realize intra-module and inter-module information interaction to achieve multimodal conversational emotion recognition. Although transformer-based methods have made progress from the perspective of global utterance sequence modeling, this paradigm underestimates the complex emotional interactions between multimodal utterances \cite{tu2024adaptive} and ignores the multiple relationships between utterances \cite{chen2023multivariate}, which limits the model's emotion recognition performance.

Benefitting from GNN's ability to mine and represent complex relationships \cite{yin2023messages, yin2024dynamic, ju2024survey}, recent GNN-based methods \cite{ai2024gcn, hu2021mmgcn, li2023ga2mif} have made significant progress in the MERC task. For instance, MMGCN \cite{hu2021mmgcn} fully connects all utterance nodes of the same modality and connects different modal nodes of the same utterance to build a heterogeneous graph to model the complex semantic relationships between multimodal utterances, then uses a deep spectral domain GNN to capture long-distance contextual information to achieve multimodal conversational emotion recognition. Although these GNN-based methods show promising performance, they still have some common limitations:

\textbf{(1) Insufficient long-distance dependence perception.} Considerable methods \cite{ghosal2019dialoguegcn, ai2024gcn, li2023graphcfc, shou2024revisiting} using sliding windows to limit the length of fully connected utterances and then using GNN to learn multimodal utterance representations to achieve emotion recognition. However, limited by the over-smoothing characteristics of GNN \cite{liu2022revisiting, yi2024fouriergnn}, usually only two layers can be stacked for capturing semantic information, making it difficult for these methods to capture long-distance emotional dependencies. Although the method \cite{hu2021mmgcn, chen2023multivariate} without a sliding window can enhance the capture of long-distance dependencies, it will cause many nodes with the non-same emotions in the neighborhood, which is not conducive to the representation learning of GNN and puts enormous performance pressure on GNN. Therefore, previous GNN-based methods still have limitations in long-distance dependency capture.

\textbf{(2) Underutilization of high-frequency features.} Many studies have shown that GNN has low-pass filtering characteristics \cite{nt2019revisiting, chang2021not, yin2022dynamic}, which mainly obtain node representation by aggregating the consistency features of the neighborhood (low-frequency information) and suppressing the dissimilarity features of the neighborhood (high-frequency information). However, consistency and dissimilarity features are equally important in the MERC task. When specific modalities express less obvious emotions, information from other modalities is needed to compensate, thereby revealing the speaker's genuine emotions. Inspired by this, M$^3$Net \cite{chen2023multivariate} tried to use high-frequency information to improve the MERC task and improved the emotion recognition effect by directly fusing high- and low-frequency features. However, essential differences exist between high and low-frequency features, and direct fusion cannot establish efficient collaboration. Thus, previous GNN-based methods still have limitations in utilizing and collaborating high and low-frequency features.

Inspired by the above analysis, to efficiently learn the consistency and complementary semantic information in multimodal conversation, we try to revisit the problem of multimodal emotion recognition in conversation from the perspective of the graph spectrum. Specifically, we propose a Graph-Spectrum-based Multimodal Consistency and Complementary feature collaboration framework GS-MCC. GS-MCC first uses RoBERTa \cite{liu2019roberta}, OpenSMILE \cite{eyben2010opensmile}, and 3D-CNN \cite{ji20123d} to extract preliminary text and acoustic and visual features. Then, GRU and a fully connected network are used further to encode text, auditory, and visual features to obtain higher-order utterance representation. In order to capture long-distance dependency information more efficiently, a sliding window is used to construct a fully connected graph to model conversational relationships, and an efficient Fourier graph operator is used to extract long-distance high and low-frequency information, respectively. In addition, to promote the collaboration ability of high and low-frequency information, we use contrastive learning to construct self-supervised signals that reflect complementarity and consistent semantic collaboration with high and low-frequency signals, thereby improving the ability of high and low-frequency information to reflect real emotions. Finally, we input the collaborative high and low-frequency information into the MLP network and softmax function for emotion prediction.

The contributions of our work are summarized as follows:
\begin{itemize}
	\item[$\bullet$] We propose an efficient long-distance information learning module that designs Fourier graph operators to build a mixed-layer GNN to capture high and low-frequency information to obtain consistency and complementary semantic dependencies in multimodal conversational contexts.

	\item[$\bullet$] We propose an efficient high- and low-frequency information collaboration module that uses contrastive learning to construct self-supervised signals that reflect the collaboration of high- and low-frequency information in terms of complementarity and consistent semantics and improves the ability to distinguish emotions between different frequency information.

	\item[$\bullet$] We conducted extensive comparative and ablation experiments on two benchmark data sets, IEMOCAP and MELD. The results show that our proposed method can efficiently capture long-distance context dependencies and improve the performance of MERC.
\end{itemize}

\section{Related work}
\label{sec:relatedwork}
Human-machine intelligent conversation systems have recently received significant attention and development \cite{lian2023gcnet, qian2023contrastive, wang2023distribution,10273224, chen2024learning}, so understanding conversations is crucial. Driven by this, Multimodal Emotion Recognition in Conversation (MERC) has gradually developed into a new research hotspot. Many researchers \cite{li2023revisiting, zhang2023dualgats, zou2023multimodal} have explored and improved the effect of MERC from the semantic interaction between text, auditory, and visual modal data in conversational contexts. These methods \cite{lin2023multi, zhang2024multi, cheng2023multimodal} agree that the MERC task focuses on better capturing and fusing multimodal semantic information in the conversational context for emotion recognition. Therefore, we will review the literature closely related to the above topics from the two aspects of multimodal conversation context feature capture and fusion.

\textbf{(1) Multimodal conversational context feature capture.} In early work, the MERC task mainly adopted GRU \cite{majumder2019dialoguernn} or LSTM \cite{poria2017context} to capture multimodal information in the conversational context. For example, \textit{Poria et al.} \cite{poria2017context} proposed a multimodal conversation emotion recognition model based on Bidirectional Long Short-Term Memory (Bi-LSTM), which captures multimodal contextual information at each time step to understand conversational context relationships in sequence data better. Although methods based on GRU or LSTM can model multimodal conversation context, they cannot capture long-distance information dependencies due to limited memory capabilities. For instance, \textit{Ma et al.} \cite{ma2023transformer} used intra-modal and inter-modal Transformers to capture semantic information in a multimodal conversation context and designed a hierarchical gating mechanism to achieve the fusion of multimodal features. Although Transformer-based methods can capture long-distance semantic information through global sequence modeling, they underestimate the complexity of multimodal dialogue semantics. Due to the superiority of GNN in modeling complex relationships, most existing research chooses to use GNN for global semantic capture and has achieved remarkable results. For example, \textit{Li et al.} \cite{li2023graphcfc} proposed directed Graph-based Cross-modal Feature Complementation (GraphCFC), which alleviates the heterogeneity gap problem in multimodal fusion by utilizing multiple subspace extractors and pairwise cross-modal complementation strategies. In addition, speaker information is vital in emotion recognition because emotions are usually subjective and individual experiences. Therefore, \textit{Ren et al.} \cite{ren2021lr} built a graph model to incorporate conversational context information and speaker dependencies, and then introduced a multi-head attention mechanism to explore potential connections between speakers.

\textbf{(2) Multimodal conversational context feature fusion.} Choosing an appropriate multimodal feature fusion strategy is another crucial step in multimodal dialogue emotion recognition \cite{chudasama2022m2fnet, zou2023multimodal}. For example, \textit{Zadeh et al.} \cite{zadeh2017tensor} proposed Tensor Fusion Network (TFN), has advantages in processing higher-order data structures (such as multi-dimensional arrays) and is therefore better able to preserve relationships between data when integrating multimodal information. So \textit{Liu et al.} \cite{liu2018efficient} proposed a Low-rank Multimodal Fusion (LMF) method. Multimodal fusion is performed using modality-specific low-order factors by decomposing tensors and weights in parallel. It avoids calculating high-dimensional tensors, reduces memory overhead, and reduces exponential time complexity to linear. \textit{Tellamekala et al.} \cite{tellamekala2023cold}  proposed Calibrated and Ordinal Latent Distribution Fusion (COLD Fusion). The proposed fusion framework involves learning the latent distribution over an unimodal temporal context by constraining the variance through calibration and ordinal ordering. Furthermore, contrastive learning has attracted increasing research attention due to its powerful ability to obtain meaningful representations through alignment fusion. \textit{Kim et al.} \cite{kim2021contrastive} introduced a contrastive loss function to facilitate impactful adversarial learning. This approach enables the adversarial learning of weak emotional samples by leveraging strong emotional samples, thereby enhancing the comprehension of intricate emotional elements embedded in intense emotions. \textit{Wang et al.} \cite{wang2022self} proposed a multimodal feature fusion framework based on contrastive learning. The framework first improves the ability to capture emotional features through contrastive learning and then uses an attention mechanism to achieve the fusion of multimodal features.

Although multimodal conversational emotion recognition has made significant progress by modeling contextual semantic information and feature fusion, the critical role of high-frequency information in MERC has been ignored. To this end, \textit{Hu et al.} \cite{hu2021mmgcn} proposed a  Multimodal Fusion Graph Convolution Network (MMGCN). MMGCN can not only capture high and low-frequency information in multimodal conversations, but also utilizes speaker information to model inter-speaker and intra-speaker dependencies. Similarly, \textit{Chen et al.} \cite{chen2023multivariate} modeled MERC from multivariate information and high- and low-frequency information, further improving the effect of multimodal conversational emotion recognition. Nevertheless, as discussed earlier, these methods do not profoundly explore the uses of high and low-frequency signals, ignoring the consistency and complementary synergy between them.

This paper starts from the perspective of graph spectrum, uses high and low-frequency signals to reconstruct MERC, captures and collaborates consistency and complementary semantic information, respectively, and improves the effect of multimodal conversational emotion recognition.

\section{Preliminary}
\subsection{Feature Extraction}

\noindent\textbf{Text Feature Extraction:} Word embeddings can capture the semantic relationships between words, making words with similar meanings closer in the embedding space. Inspired by previous work \cite{kim2021emoberta, chudasama2022m2fnet, shen2021directed}, we use the RoBERTa model \cite{liu2019roberta} to extract text features and the embedding is denoted as \({\varphi}_{{t}}\).

\noindent\textbf{Audio and Vision Feature Extraction:} Consistent with previous work \cite{li2022emocaps,ghosal2019dialoguegcn,majumder2019dialoguernn}, we employ openSMILE and 3D-CNN for audio and Vision feature extraction, yielding respective embeddings \({\varphi}_{{a}}\) and \({\varphi}_{{v}}\).

\subsection{Speaker information embedding}
Speaker information can play an important role in emotion recognition. Emotion is not only related to the characteristic attributes of the utterance but also to the speaker's inherent expression manner. Inspired by previous work \cite{hu2021mmgcn, chen2023multivariate, zhang2024multi}, we incorporate speaker information into each unimodal utterance to obtain an unimodal representation of context and speaker information.

\begin{figure*}[htbp]
	\centering
	\includegraphics[width=1.0\linewidth]{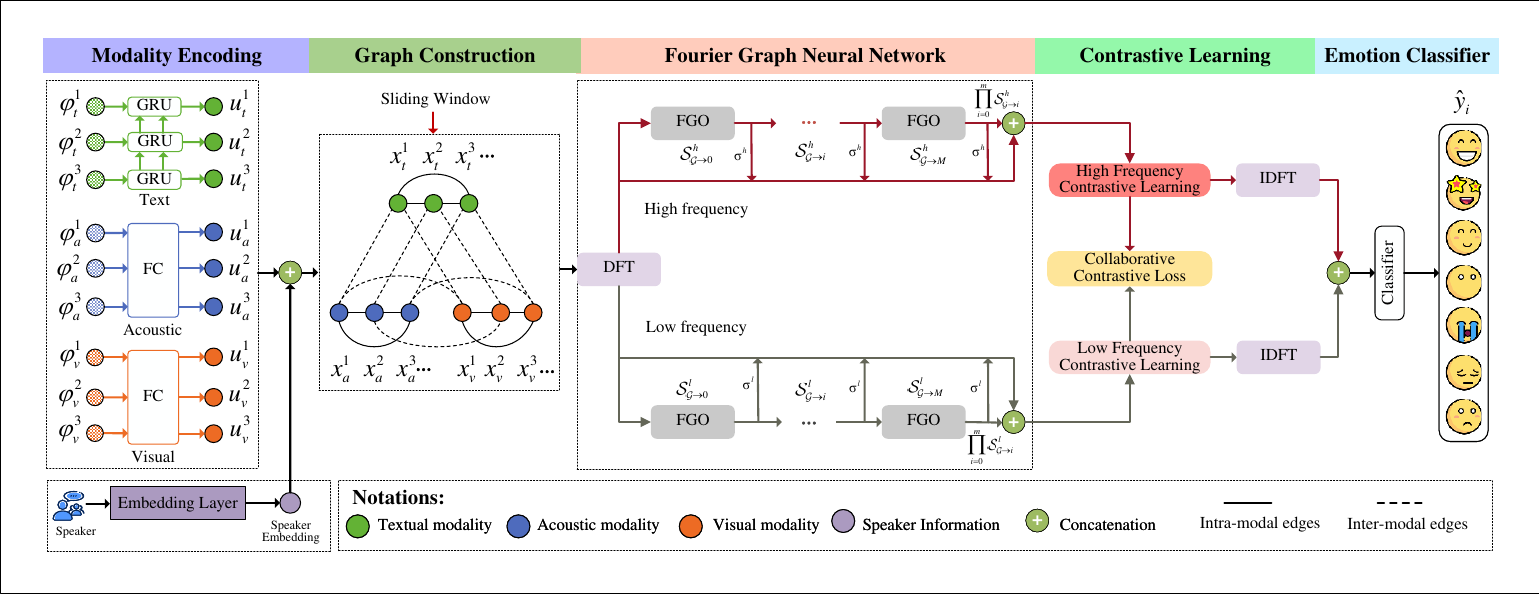}
	\caption{The overall architecture of the proposed model GS-MCC. Specifically, feature embedding of multimodal utterances and speaker information is first performed, and then the embedded features are used to construct a multimodal semantic interaction graph. Then, a Fourier graph neural network is used to capture long-distance dependent high and low-frequency information, and finally, contrastive learning is used to collaborate high and low-frequency information for emotion recognition.}
	\label{fig_2}
\end{figure*}

Specifically, we first sort all speakers by name and then use the one-hot vector $s_{i}$ to represent the $i$-th speaker. Finally, we perform a unified embedding representation for the speakers to make similar speakers closer together in the embedding space. The embedding of the $i$-th speaker is as follows:
\begin{equation}
	S_{i}=W_{speaker}s_{i},
\end{equation}
where $W_{speaker}$ is the trainable weight. In addition, to obtain higher-order feature representation, we utilize bidirectional Gated Recurrent Units (GRU) to encode conversational text features. We have observed in practice that using recursive modules to encode visual and auditory modalities has no positive performance impact. Therefore, we employed a multilayer perceptron with two single hidden layers to encode auditory and visual modalities, respectively. The specific encoding calculation is as follows:
\begin{equation}
	\begin{aligned}
		&u_{t} =\overleftrightarrow{GRU}({\varphi}_{{t}},u_t^{(+,-)1}),  \\
		&u_{a} =W_a{\varphi}_{{a}}+b_a,  \\
		&u_{v} =W_v{\varphi}_{{v}}+b_v,
	\end{aligned}
\end{equation}
where ${W}_{a}$, $b_a$, ${W}_{v}$ and $b_v$ are the learnable parameters of the auditory and visual encoders, respectively. We then add speaker embeddings to obtain speaker- and context-aware unimodal representations:
\begin{equation}
	x_{m}=u_{m}+S_{i},\quad m\in\{t,a,v\},
\end{equation}
where $t, a, v$ represent text, audio, and vision modal, respectively.

\section{Methodology}
Fig. \ref{fig_2} shows the proposed Graph-Spectrum-based Multimodal Consistency and Complementary collaborative learning framework GS-MCC. GS-MCC contains five modules: feature encoding, multimodal interaction graph construction, Fourier graph neural network, contrastive learning, and emotion classification.

\subsection{Multimodal Interaction Graph}
To model the latent semantic dependencies between multimodal utterances, we adopt a multimodal interaction graph for construction. Instead of fully connecting all nodes of the same modality, we use a sliding window for restriction. Although fully connecting all nodes of the same modality is beneficial to building long-distance semantic dependencies, it will introduce much noise, which is not conducive to subsequent GNN learning.

Given a conversation sequence $U = \left\{ {{u_1},...,{u_N}} \right\}$ with $N$ multimodal utterances, under the restriction of the sliding window $k$, we can construct a multimodal interaction graph ${G^k} = \left( {{V^k},{E^k},{A^k},{X^k}} \right)$, where the node $v \in {V^k}$ represents a single-modal utterance and the edge $e \in {E^k}$ represents two semantic interactive relationships between unimodal utterances, ${A^k}$ is the adjacency matrix, and ${X^k}$ is the feature matrix. The multimodal semantic interaction graph is constructed as follows:

\textbf{Nodes:} Since any utterance ${u_i} \in U$ contains three modal information, we treat each modality in each utterance as an independent node, using text modal node $x_t^i$, auditory modal node $x_a^i$, and visual modal node $x_v^i$ represents, and uses the corresponding features $x_m^i$ to represent the initial embedding of the node. The constructed multimodal interaction graph ${G^k}$ has 3$N$ nodes.

\textbf{Edges:} In order to avoid introducing noise or redundant information, we use a sliding window to limit node connections of the same mode. Specifically, we fully connect the nodes in the same mode within sliding window $k$. In addition, we connect different modal nodes of the same utterance to construct semantic interactions between modalities. For example, for utterance ${u_i} \in U$, connections need to be constructed between nodes $x_t^i$, $x_a^i$, and $x_v^i$ in different modalities.

\textbf{Edge Weight Initialization:} In order to better capture the similarity between nodes, we use different similarities to determine edge weights for different types of edges. Nodes with higher similarity show more critical information interactions between them. Specifically, for edges coming from nodes of the same modality, since the feature distribution of the nodes is potentially consistent, our calculation method is as follows:
\begin{equation}
	{A}_{ij}^k=1 -\frac{\arccos\left(sim(x_m^i,x_m^j)\right)}{\pi},
\end{equation}
where $x_m^i$ and $x_m^j$ represent the feature representations of the $i$-th and $j$-th nodes in the graph. For edges between nodes in different modalities, since the feature distribution of the nodes is not potentially consistent, we use the hyperparameter $\phi$ to optimize the similarity learning between cross-modal nodes. Our approach is computed as follows:
\begin{equation}
	{A}_{ij}^k=\phi\left(1-\frac{\arccos\left(sim(x_m^i,x_m^j)\right)}{\pi}\right).
\end{equation}

\subsection{Fourier Graph Neural Network}
As mentioned above, using a sliding window will limit long-distance dependency learning. This is because traditional GNN has over-smoothing characteristics and cannot stack many layers. Different from the methods used by MMGCN \cite{hu2021mmgcn} and M$^3$Net \cite{chen2023multivariate}, this paper is inspired by FourierGNN \cite{yi2024fouriergnn}, designs efficient Fourier graph operators for high and low-frequency signals, respectively, to capture the long-distance dependency information.

\textbf{Fourier Graph Operator.} For a given multimodal interaction graph, ${G^k} = \left( {{V^k},{E^k},{A^k},{X^k}} \right)$, where ${A^k} \in {\mathbb{R}^{3N \times 3N}}$ is the adjacency matrix, ${X^k} \in {\mathbb{R}^{3N \times d}}$ is the feature matrix, $N$ is the number of multimodal utterances, and $d$ is the dimension of the feature. According to FourierGNN, we can obtain the Green kernel $\kappa  \in {\mathbb{R}^{d \times d}}$ that meets the conditions based on the adjacency matrix $A^k$ and the weight matrix $W \in {\mathbb{R}^{d \times d}}$, which needs to satisfy the conditions $\kappa \left[ {i,j} \right] = \kappa \left[ {i - j} \right]$, $\kappa \left[ {i,j} \right] = {A_{ij}^k} \circ W$, and $i$ and $j$ are fall between 1 and 3$N$. Based on the kernel $\kappa$, we can obtain the following Fourier graph operator ${{\mathcal S}_{\mathcal G}}$:
\begin{equation}
	{{\mathcal S}_{\mathcal G}} = {\mathcal F}\left( \kappa  \right) \in {\mathbb{C}^{^{3N \times d \times d}}},
\label{form5}
\end{equation}
where $\mathcal{F}$ is the Discrete Fourier Transform (DFT). According to the graph convolution theory, we can express the graph convolution operation as follows:
\begin{equation}
	{F_{\theta _{\mathcal{G}}}}\left( {{X^k},{A^k}} \right) = {A^k}{X^k}W = {{\mathcal{F}}^{ - 1}}\left( {{\mathcal{F}}\left( {X^k} \right){\mathcal{F}}\left( \kappa  \right)} \right),
\label{form6}
\end{equation}
where ${\theta _{\mathcal G}}$ is the learnable parameter and ${{\mathcal{F}}^{ - 1}}$ is the Inverse Discrete Fourier Transform (IDFT). According to the convolution theory and the conditions of FGO, we can expand the frequency domain term in Eq. (\ref{form6}) as follows:
\begin{equation}
    \begin{aligned}
	{\mathcal F}\left( {{X^k}} \right){\mathcal F}\left( {\kappa} \right) &= {\mathcal F}\left( {\left( {{X^k}*{\kappa}} \right)\left[ i \right]} \right) \\
&= {\mathcal F}\left( {{X^k}\left[ j \right]{\kappa}\left[ {i - j} \right]} \right) = {\mathcal F}\left( {{X^k}\left[ j \right]{\kappa}\left[ {i,j} \right]} \right) \\
&= {\mathcal F}\left( {A_{ij}^k{X^k}\left[ j \right]W} \right) = {\mathcal F}\left( {{A^k}{X^k}W} \right).
\end{aligned}
\label{form7}
\end{equation}

As seen from Eq. (\ref{form7}) , the graph convolution operation is implemented through the product of FGO and features in the frequency domain. In addition, according to the convolution theory, the convolution of time-domain signals is equal to the product of frequency-domain signals. The product operation in the frequency domain only requires $O$($N$) time complexity, while the convolution operation in the time domain requires $O\left( {{N^2}} \right)$ time complexity. Therefore, an efficient graph neural network can be constructed based on the Fourier graph operator.

To efficiently capture high- and low-frequency information, we perform targeted optimization on FGO and use the high-pass and low-pass filters to extract complementary and consistent semantic information. The specific filter design is as follows:
\begin{equation}
	{L^l} = I + D_{\mathcal G}^{ - 1/2}{A^k}D_{\mathcal G}^{ - 1/2},
\label{form8}
\end{equation}
\begin{equation}
	{L^h} = I - D_{\mathcal G}^{ - 1/2}{A^k}D_{\mathcal G}^{ - 1/2},
\label{form9}
\end{equation}
where $I$ is the identity matrix, ${D_{\mathcal G}}$ and ${A^T}$ are the degree matrix and adjacency matrix of the multimodal interaction graph, respectively, and ${L^l}$ and ${L^h}$ are the low-pass and high-pass filters, respectively. Based on low-pass and high-pass filters, we can obtain the following low and high-frequency Green kernel and Fourier graph operator:
\begin{equation}
	{\kappa ^{l/h}}\left[ {i,j} \right] = L_{ij}^{l/h} \circ W,
\label{form10}
\end{equation}
\begin{equation}
	{\mathcal S}_{\mathcal G}^{l/h} = {\mathcal F}\left( {{\kappa ^{l/h}}} \right).
\label{form11}
\end{equation}

Finally, we can build an $M$-layer Fourier graph neural network based on these efficient Fourier graph operators to capture long-distance high and low-frequency dependency information in multimodal interaction graphs:
\begin{equation}
	F_{{\theta _{\mathcal G}}}^{l/h}\left( {{X^k},{A^k}} \right) = \sum\limits_{m = 0}^M \sigma  \left( {{\mathcal F}({X^k}){\mathcal S}_{{\mathcal G} \Rightarrow \left[ {0:m} \right]}^{l/h} + {b_{l/h}}} \right),
\label{form12}
\end{equation}
\begin{equation}
	{\mathcal S}_{{\mathcal G} \Rightarrow \left[ {0:m} \right]}^{l/h} = \prod\limits_{i = 0}^m {{\mathcal S}_{{\mathcal G} \to i}^{l/h}},
\label{form13}
\end{equation}
where $\sigma $ is the activation function, ${{b_{l/h}}}$ is the bias parameter, ${{\mathcal S}_{{\mathcal G} \to i}^{l/h}}$ is the FGO in the $i$-th layer, $l$, and $h$ represent low and high frequencies respectively.

By stacking $M$ layers of Fourier graph operators, our model can capture long-distance dependency information and obtain each node's low-frequency feature representation, $x_m^l$, and high-frequency feature representation, $x_m^h$, respectively.

\subsection{Contrastive Learning}
Low-frequency features reflect the trend of slow changes in emotion, while high-frequency features reflect the trend of rapid changes in emotion. To synergize these two features, we employ contrastive learning to build self-supervised signals to promote consistent and complementary semantics learning in multimodal utterances.

Inspired by the SpCo \cite{liu2022revisiting} method, increasing the frequency domain difference between two contrasting views can achieve better contrast learning effects. Unlike SpCo, our contrastive learning is performed directly in the frequency domain and does not rely on data augmentation to generate contrastive views. Specifically, we use a combination of low-frequency contrast learning and high-frequency contrast learning to promote the synergy of the two features. In addition, we only use the strategy of negative sample pairs far away from each other to increase the frequency domain difference between contrasting views and obtain better contrast learning effects.

\textbf{LFCL: Low Frequency Contrastive Learning.} LFCL aims to use low-frequency samples as anchor nodes and all high-frequency nodes as negative samples to construct a self-supervised signal to increase the frequency domain difference between contrast views to obtain better contrast learning effects and promote consistent semantics and complementary semantics learning in multimodal conversations. For each low-frequency anchor node, the self-supervised contrast loss can be defined as:
\begin{equation}
	{{\mathcal L}_{IFCL}} =  - \frac{1}{\tau } + \log \left( {{e^{1/\tau }} + \sum\limits_{i = 1}^{3N} {{e^{\left( {{{\left( {x_m^l} \right)}^T}x_m^{hi - }} \right)/\tau }}} } \right),
	\label{form14}
\end{equation}
where $\tau $ is the temperature coefficient, ${x_m^l}$ is the low-frequency anchor node, and ${x_m^{hi - }}$ is the $i$-th high-frequency negative sample.

\textbf{HFCL: High Frequency Contrastive Learning.} HFCL is similar to LFCL, except that HFCL uses high-frequency samples as anchor nodes and all low-frequency nodes as negative samples to construct a self-supervised signal to increase the frequency domain difference between contrasting views. The specific contrast loss can be defined as:
\begin{equation}
	{{\mathcal L}_{HFCL}} =  - \frac{1}{\tau } + \log \left( {{e^{1/\tau }} + \sum\limits_{i = 1}^{3N} {{e^{\left( {{{\left( {x_m^h} \right)}^T}x_m^{li - }} \right)/\tau }}} } \right),
	\label{form15}
\end{equation}
where ${x_m^h}$ is the high-frequency anchor node, and ${x_m^{li - }}$ is the $i$-th low-frequency negative sample.

The overall collaborative contrastive learning loss is the sum of LFCL and HFCL, which can be expressed as ${\mathcal L}_{CCL}$:
\begin{equation}
	{{\mathcal L}_{CCL}} = {{\mathcal L}_{LFCL}} + {{\mathcal L}_{HFCL}}.
	\label{form16}
\end{equation}

Finally, we use the inverse discrete Fourier transform to convert the high and low-frequency features into time domain features and concatenation the two parts of features to obtain the final embedding representation of the uni-modal utterance node:
\begin{equation}
	{v_m}{\rm{ = IDFT}}\left( {x_m^l} \right) \oplus {\rm{IDFT}}\left( {x_m^h} \right),
	\label{form17}
\end{equation}
where $m\in\{t,a,v\}$ represents any one of text, auditory and visual modalities.

\subsection{Emotion Classifier}
For modal utterance $U_i$, we concatenate the features of each modality for emotion classification.
\begin{equation}
	{U_i} = v_t^i \oplus v_a^i \oplus v_v^i,
	\label{form18}
\end{equation}
\begin{equation}
	\tilde{U}_{i} =\operatorname{ReLU}(U_i),
	\label{form19}
\end{equation}
\begin{equation}
	\mathcal{P}_{i} =\operatorname{softmax}(W_u\tilde{U}_i+b_u),
	\label{form20}
\end{equation}
\begin{equation}
	\hat{y}_{i} =\operatorname{argmax}(\mathcal{P}_i[\tau]),
	\label{form21}
\end{equation}
where $W_u$ and $b_u$ are learnable parameters, and $\hat{y}_i$ is the predicted emotion label of utterance $U_i$. Finally, we employ categorical cross-entropy loss and contrastive loss for model training.

\section{Experiments}
\subsection{Implementation Details}
\textbf{Benchmark Datasets and Evaluation Metrics:}
In our experiments, we used two multimodal datasets, IEMOCAP \cite{busso2008iemocap}, and MELD \cite{poria2019meld}, widely used in multimodal emotion recognition.  IEMOCAP (Interactive Emotional Dyadic Motion Capture Database) is a multimodal database for emotion recognition and analysis. The IEMOCAP data set consists of movie dialogue clips and emotional annotations, including voice, video, and emotional annotation data of 10 actors in interactive dialogue scenes. MELD (Multimodal EmotionLines Dataset) contains dialogue text from movie and TV show clips. The dialogue text contains the characters' speech and the context information of the dialogue. MELD also provides audio recordings and video recordings of conversations. We record the classification accuracy (Acc.) and F1 for each emotion category, as well as the overall weighted average accuracy (W-Acc.) and weighted average F1 (W-F1).

\noindent \textbf{Baseline Methods:}
We compare several baselines on the IEMOCAP and MELD datasets, including bc-LSTM \cite{poria2017context}, and A-DMN \cite{xing2020adapted} based on RNN architecture, LFM \cite{liu2018efficient} based on Low-rank Tensor Fusion network, DialogueGCN \cite{ghosal2019dialoguegcn}, LR-GCN \cite{ren2021lr}, DER-GCN \cite{ai2024gcn}, MMGCN \cite{hu2021mmgcn}, AdaGIN \cite{tu2024adaptive}, RGAT \cite{ishiwatari2020relation} and CoMPM \cite{lee2022compm} based on GCN, EmoBERTa \cite{kim2021emoberta},  CTNet \cite{lian2021ctnet} and COGMEN \cite{joshi2022cogmen} based on Transformer architecture.

\noindent \textbf{Experimental Setup:}
All experiments are conducted using Python 3.8 and PyTorch 1.8 deep learning framework and performed on a single NVIDIA RTX 3090 24G GPU. Our model is trained using AdamW with a learning rate of 1e-5, cross-entropy as the loss function, and a batch size of 32. The optimal parameters of all models were obtained by performing parameter adjustment using the leave-one-out cross-validation method on the validation set.

\begin{table*}[htbp]
	\setlength{\tabcolsep}{6pt}
	\centering
	\caption{Comparison with other baseline models on the IEMOCAP dataset. The best result in each column is in bold.}
	\label{tab:iemocap}
	\begin{tabular}{@{}l|c|cccccccccccccc@{}}
		\toprule
		\multirow{3}{*}{Methods} & \multirow{3}{*}{Parmas.} & \multicolumn{14}{c}{IEMOCAP}                                                                                                                                                                                  \\ \cmidrule(l){3-16}
		&                          & \multicolumn{2}{c}{Happy} & \multicolumn{2}{c}{Sad} & \multicolumn{2}{c}{Neutral} & \multicolumn{2}{c}{Angry} & \multicolumn{2}{c}{Excited} & \multicolumn{2}{c}{Frustrated} & \multicolumn{2}{c}{Average(w)} \\
		&                          & Acc.        & F1          & Acc.       & F1         & Acc.         & F1           & Acc.        & F1          & Acc.         & F1           & Acc.           & F1            & Acc.           & F1            \\ \midrule
		bc-LSTM                  &    1.28M                      & 29.1        & 34.4        & 57.1       & 60.8       & 54.1         & 51.8         & 57.0        & 56.7        & 51.1         & 57.9         & 67.1           & 58.9          & 55.2           & 54.9          \\
		LFM                      &   2.34M                       & 25.6        & 33.1        & 75.1       & 78.8       & 58.5         & 59.2         & 64.7        & 65.2        & 80.2         & 71.8         & 61.1           & 58.9          & 63.4           & 62.7          \\
		A-DMN                    &       --                   & 43.1        & 50.6        & 69.4       & 76.8       & 63.0         & 62.9         & 63.5        & 56.5        & \textbf{88.3}         & 77.9         & 53.3           & 55.7          & 64.6           & 64.3          \\
		DialogueGCN              &     12.92M                     & 40.6        & 42.7        & \textbf{89.1}       & \textbf{84.5}       & 62.0         & 63.5         & 67.5        & 64.1        & 65.5         & 63.1         & 64.1           & 66.9          & 65.2           & 64.1          \\
		RGAT                     &    15.28M                      & 60.1        & 51.6        & 78.8       & 77.3       & 60.1         & 65.4         & 70.7        & 63.0        & 78.0         & 68.0         & 64.3           & 61.2          & 65.0           & 65.2          \\
		CoMPM                    &       --                   & 59.9        & 60.7        & 78.0       & 82.2       & 60.4         & 63.0         & 70.2        & 59.9        & 85.8         & 78.2         & 62.9           & 59.5          & 67.7           & 67.2          \\
		EmoBERTa                 & 499M                         & 56.9        & 56.4        & 79.1       & 83.0       & 64.0         & 61.5         & 70.6        & 69.6        & 86.0         & 78.0         & 63.8           & 68.7          & 67.3           & 67.3          \\
		COGMEN                   &   --                       & 57.4        & 51.9        & 81.4       & 81.7       & 65.4         & 68.6         & 69.5        & 66.0        & 83.3         & 75.3         & 63.8           & 68.2          & 68.2           & 67.6          \\
		CTNet                    &    8.49M                      & 47.9        & 51.3        & 78.0       & 79.9       & 69.0         & 65.8         & \textbf{72.9}        & 67.2        & 85.3         & \textbf{78.7}         & 52.2           & 58.8          & 68.0           & 67.5          \\
		LR-GCN                   &    15.77M                      & 54.2        & 55.5        & 81.6       & 79.1       & 59.1         & 63.8         & 69.4        & 69.0        & 76.3         & 74.0         & 68.2           & 68.9          & 68.5           & 68.3          \\
		MMGCN  & 0.46M  &  43.1 &  42.3   &  79.3     &  78.7 &  63.5  & 61.7   &  69.6    &  69.0 &  75.8  &    74.3    & 63.5  & 62.3  &  67.4  & 66.2 \\
		AdaGIN       & 6.3M  & 53.0        &   --    & 81.5    &  --  &  71.3   &  -- & 65.9   & -- &  76.3  &  -- & 67.8 &  -- & 70.5  & 70.7  \\
		DER-GCN                  &  78.59M                        & \textbf{60.7}        & 58.8        & 75.9       & 79.8       & 66.5         & 61.5         & 71.3        & \textbf{72.1}        & 71.1         & 73.3         & 66.1           & 67.8          & 69.7           & 69.4         \\  \hdashline
		GS-MCC & 2.10M & 60.2 & \textbf{65.4} & 86.2 & 81.2 & \textbf{75.7} & \textbf{70.9} & 71.7 & 70.8 & 83.2 & 81.4 & \textbf{66.0} & \textbf{71.0} & \textbf{73.8} & \textbf{73.9} \\ \bottomrule
	\end{tabular}
\end{table*}


\begin{table*}[htbp]
	\caption{Comparison with other baseline models on the MELD dataset. The best result in each column is in bold.}
	\label{tab:meld}
	\setlength{\tabcolsep}{5pt}{
		\begin{tabular}{@{}l|c|cccccccccccccccl@{}}
			\toprule
			\multirow{3}{*}{Methods} & \multirow{3}{*}{Parmas.} & \multicolumn{16}{c}{MELD}                                                                                                                                                                                                             \\ \cmidrule(l){3-18}
			&                          & \multicolumn{2}{c}{Neutral} & \multicolumn{2}{c}{Surprise} & \multicolumn{2}{c}{Fear} & \multicolumn{2}{c}{Sadness} & \multicolumn{2}{c}{Joy} & \multicolumn{2}{c}{Disgust} & \multicolumn{2}{c}{Anger} & \multicolumn{2}{c}{Average(w)} \\
			&                          & Acc.         & F1           & Acc.          & F1           & Acc.        & F1         & Acc.         & F1           & Acc.       & F1         & Acc.         & F1           & Acc.        & F1          & Acc.  & \multicolumn{1}{c}{F1} \\ \midrule
			bc-LSTM                  &    1.28M                      & 78.4         & 73.8         & 46.8          & 47.7         & 3.8         & 5.4        & 22.4         & 25.1         & 51.6       & 51.3       & 4.3          & 5.2          & 36.7        & 38.4        & 57.5  & 55.9                   \\
			DialogueRNN              &  14.47M                        & 72.1         & 73.5         & 54.4          & 49.4         & 1.6         & 1.2        & 23.9         & 23.8         & 52.0       & 50.7       & 1.5          & 1.7          & 41.0        & 41.5        & 56.1  & 55.9                   \\
			DialogueGCN              &  12.92M                        & 70.3         & 72.1         & 42.4          & 41.7         & 3.0         & 2.8        & 20.9         & 21.8         & 44.7       & 44.2       & 6.5          & 6.7          & 39.0        & 36.5        & 54.9  & 54.7                   \\
			RGAT                     &  15.28M                        & 76.0         & 78.1         & 40.1          & 41.5         & 3.0         & 2.4        & 32.1         & 30.7         & 68.1       & 58.6       & 4.5          & 2.2          & 40.0        & 44.6        & 60.3  & 61.1                   \\
			CoMPM                    &   --                       & 78.3         & 82.0         & 48.3          & 49.2         & 1.7         & 2.9        & 35.9         & 32.3         & 71.4       & 61.5       & 3.1          & 2.8          & 42.2        & 45.8        & 64.1  & 65.3                   \\
			EmoBERTa                 &   499M                       & \textbf{78.9}         & \textbf{82.5}         & 50.2          & 50.2         & 1.8         & 1.9        & 33.3         & 31.2         & 72.1       & 61.7       & 9.1          & 2.5          & 43.3        & 46.4        & 64.1  & 65.2                   \\
			A-DMN                    &      --                    & 76.5         & 78.9         & 56.2          & 55.3         & 8.2         & 8.6        & 22.1         & 24.9         & 59.8       & 57.4       & 1.2          & 3.4          & 41.3        & 40.9        & 61.5  & 60.4                   \\
			LR-GCN                   &     15.77M                     & 76.7         & 80.0         & 53.3          & 55.2         & 0.0         & 0.0        & 49.6         & 35.1         & 68.0       & \textbf{64.4}       & 10.7         & 2.7          & 48.0        & 51.0        & 65.7  & 65.6                   \\
			MM-GCN                   &     0.46M                     &   64.8      &             77.1     &    67.4     &    53.9      &    0.0     &   0.0       &     72.4     &    17.7    &   68.7    &    56.9      &     0.0       &   0.0    &     54.4    & 42.6  & 64.4 &   59.4               \\
			AdaGIN       & 6.3M  & 79.8        &   --    & 60.5    & --   &  15.2   & --  & 43.7   & -- &  64.5  & --  & 29.3 & -- & 56.2 &  -- & 67.6  & 66.8  \\
			DER-GCN                  &   78.59M                       & 76.8         & 80.6         & 50.5          & 51.0         & 14.8        & \textbf{10.4}       & \textbf{56.7}         & 41.5         & 69.3       & 64.3       & 17.2         & 10.3         & 52.5        & \textbf{57.4}        & 66.8  & 66.1     \\ \hdashline
			GS-MCC & 2.10M  & 78.4 & 81.8 & \textbf{56.9}  & \textbf{58.3}   & \textbf{23.5}  & 23.8   & 50.0   & 35.8   & 69.4  & 66.4  & \textbf{36.7}    & \textbf{30.7}  & \textbf{53.2}  & 54.4 & \textbf{68.1} &  \textbf{69.0} \\ \bottomrule
	\end{tabular}}
\end{table*}

\subsection{Comparison with the State-of-the-Art}
Table \ref{tab:iemocap} and Table \ref{tab:meld} show the emotion recognition effects of the proposed GS-MCC method and the baseline method on the IEMOCAP and MELD datasets, respectively. On the IEMOCAP dataset, GS-MCC has the best emotion recognition effect, outperforming all comparison baselines, and is 3.3\% and 3.2\% better than AdaGIN on W-Acc and W-F1 ,respectively. In addition, GS-MCC has significantly improved Acc and F1 values in some emotion categories. Similarly, compared with all comparison baselines on the MELD data set, GS-MCC also has the best emotion recognition effect, outperforming AdaGIN by 0.5\% and 2.2\% on W-Acc and W-F1, respectively.  Furthermore, AdaGIN is optimal in both Acc and F1  most emotion categories. Experimental results demonstrate the effectiveness of AdaGIN.  The performance improvement may be attributed to the proposed method's ability to utilize long-distance contextual semantic information from fully- and low-frequency signals while avoiding the over-smoothing phenomenon of GCN.

Furthermore, the proposed GS-MCC has only 2.10M model parameters, which is far lower than DialogueGCN and other GCN-based emotion recognition methods. Experimental results also demonstrate the potential application of our method in efficient computing.

\begin{figure}
	\centering
	\includegraphics[width=1.0\linewidth]{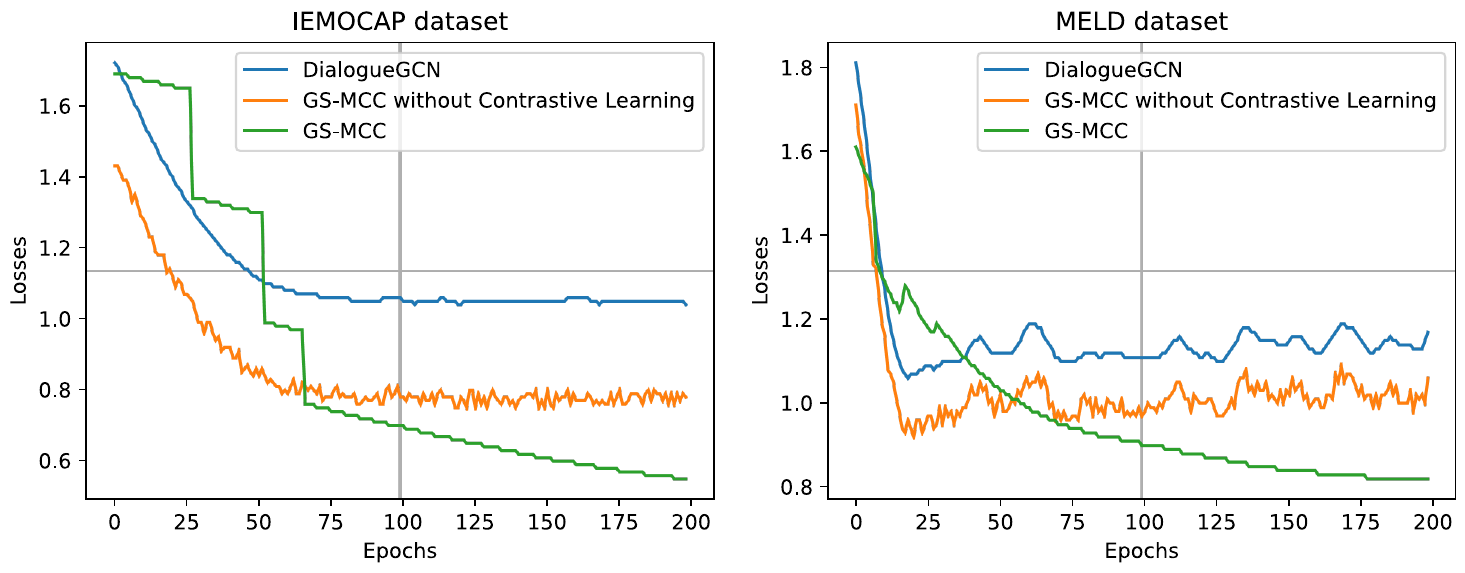}
	\caption{Loss trends during model training and inference on the IEMOCAP and MELD datasets. We compare DialogueGCN, GS-MCC without contrastive loss and GS-MCC.}
	\label{fig:fourierloss}
\end{figure}

\subsection{Trends of Losses}
During the training and inference process of the model, we show the loss trends of DialogueGCN, GS-MCC without contrastive loss, and GS-MCC in the IEMOCAP and MELD datasets to better understand the convergence of the model. Fig. \ref{fig:fourierloss} shows the results of the training loss. On the IEMOCAP data set, we found that DialogueGCN quickly converged to the local optimal value and continued fluctuating around the loss value of 1.1. The convergence of GS-MCC without contrastive loss is better than DialogueGCN, and it converges around a loss value of 0.8. Although the loss value of GS-MCC without contrastive loss is higher than GS-MCC at the beginning of training, as the training continues, the convergence of GS-MCC begins to be better than GS-MCC without contrastive loss. It converges around the loss value of 0.4. The MELD dataset's loss values of DialogueGCN and GS-MCC without contrastive loss fluctuate considerably. They are difficult to converge, but the loss of GS-MCC without contrastive loss is lower than DialogueGCN. However, GS-MCC with contrastive loss has better convergence, converging around a loss value of 0.9. Experimental results prove that the contrastive learning mechanism plays an essential role in the convergence of the GCN network and can collaborate with high- and low-frequency features for better emotion recognition.

\subsection{Over-smoothing Analysis}
\begin{figure}
	\centering
	\includegraphics[width=1\linewidth]{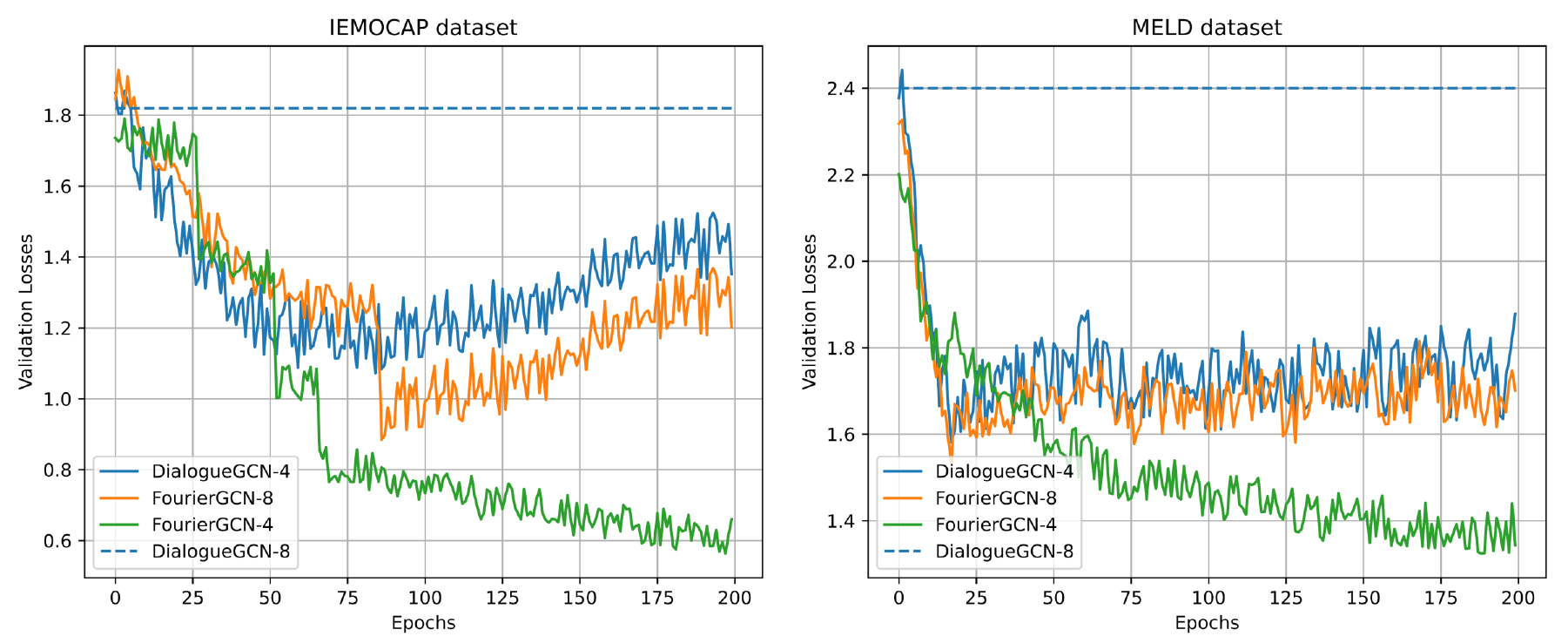}
	\caption{Emotion recognition performance of DialogueGCN and GS-MCC on IEMOCAP and MELD datasets. We stack 4-layer and 8-layer GCN to explore the over-smoothing phenomenon of the model.}
	\label{fig:oversmooth1}
\end{figure}

It is challenging to train a deep GCN with strong feature expression ability because deep GCN is prone to over-smoothing, which limits the feature expression ability of nodes. From a training perspective, over-smoothing removes discriminative semantic information from node features. Therefore, we stack 4-layer and 8-layer GCN to explore the over-smoothing phenomenon of the model's DialogueGCN and GS-MCC on the IEMOCAP and MELD datasets. Fig \ref{fig:oversmooth1} shows the experimental comparison results. On the IEMOCAP and MELD datasets, we observed that the training convergence of DailogueGCN-8 was poor, and the model suffered from severe over-smoothing. The training convergence of DailogueGCN-4 is slightly better than that of DailogueGCN-8, but it can only fluctuate around the local optimal value. DailogueGCN-4 also suffers from serious overfitting, especially on the IEMOCAP data set. Compared with DailogueGCN-8, GS-MCC-8 can alleviate the over-smoothing problem to a certain extent and converge to a local optimal value. The convergence of GS-MCC-4 is perfect and can converge to a relatively stable optimal solution. Experimental results show that GS-MCC can alleviate the model's over-smoothing problem to a certain extent. This may be attributed to GS-MCC's ability to use different-order node information of nodes in the graph to update the feature representation of nodes. By mixing the feature information of nodes of different orders in each layer, GS-MCC can maintain the diversity of node features, thereby preventing over-smoothing of features. Therefore, GS-MCC can effectively capture long-distance dependency information in multi-modal conversations.

\subsection{Ablation Study}
\textbf{Ablation studies for SE, Fourier GNN, CL.} Speaker embedding (SE), Fourier graph neural network (Fourier GNN), and contrastive learning (CL) are the three critical components of our proposed multimodal emotion recognition model. We only remove one proposed module at a time to verify the effectiveness of the component. It is worth noting that when Fourier GNN is removed, we use DialogueGCN as the backbone of the model. From the emotion recognition results in Table \ref{tab:abla}, we conclude: (1) All the modules we proposed are helpful because no matter which proposed module is deleted, it will cause the emotion recognition performance of the model to decrease. (2) Speaker embedding has a relatively significant impact on the emotion recognition performance of the model because if the speaker embedding information is removed from the IEMOCAP and MELD data sets, the emotion recognition effect of the model will be significantly reduced. The experimental results show that the speaker's embedded information is essential for the model to understand emotions. (3) On the IEMOCAP and MELD datasets, Fourier GNN is more critical than contrastive learning. We speculate that this is because Fourier GNN can capture high and low-frequency signals to provide more useful emotional semantic information, and the contrastive learning mechanism mainly assists Fourier GNN in better achieving complementary and consistent semantic information collaboration.

\begin{table}[htbp]
	\centering
	\caption{Ablation studies for SE, Fourier GNN, CL on the IEMOCAP and MELD datasets.}
	\label{tab:abla}
	\setlength{\tabcolsep}{6.7pt}{
		\begin{tabular}{@{}l|llll@{}}
			\toprule
			\multirow{2}{*}{Methods} & \multicolumn{2}{c}{IEMOCAP} & \multicolumn{2}{c}{MELD} \\ \cmidrule(l){2-5}
			& W-Acc.         & W-F1         & W-Acc.        & W-F1       \\ \midrule
			GS-MCC                     &    \textbf{73.1 }         &  \textbf{73.3 }           &  \textbf{ 68.1  }        &  \textbf{ 69.0 }        \\ \hdashline
			w/o SE                   &  $70.3_{(\downarrow 2.8)}$           &  $70.6_{(\downarrow 2.7)}$         &    $65.4_{(\downarrow 2.7)}$         &    $64.6_{(\downarrow 4.4)}$        \\
			w/o Fourier GCN                  &   $68.7_{(\downarrow 4.4)}$          & $  67.7_{(\downarrow 5.6)}$        &  $64.2_{(\downarrow 3.9)}$           &   $64.1_{(\downarrow 4.9)}$         \\
			w/o CL                  &  $ 70.3_{(\downarrow 2.8)}$          &  $71.3_{(\downarrow 2.0)}$          &    $66.1_{(\downarrow 2.0)}$         & $65.9_{(\downarrow 3.1)}$           \\ \bottomrule
	\end{tabular}}
\end{table}

\begin{table}[htbp]
	\centering
	\setlength{\tabcolsep}{2.5mm}{
		\caption{The effect of our method on IEMOCAP and MELD datasets using unimodal features and multimodal features, respectively.}
		\label{tab:multimodal}
		\begin{tabular}{@{}c|llll@{}}
			\toprule
			\multirow{2}{*}{Modality} & \multicolumn{2}{c}{IEMOCAP} & \multicolumn{2}{c}{MELD} \\ \cmidrule{2-5}
			& W-Acc.       & W-F1              & W-Acc.         & W-F1         \\ \hline
			T+A+V                     &\textbf{73.8}      & \textbf{73.9}               &    \textbf{68.1}        &  \textbf{69.0} \\ 
			T                         & $66.3_{(\downarrow 7.5)}$        & $66.0_{(\downarrow 7.9)}$       & $63.7_{(\downarrow 4.4)}$          & $62.5_{(\downarrow 6.5)}$           \\
			A                         & $57.7_{(\downarrow 16.1)}$        & $58.1_{(\downarrow 15.8)}$                & $53.8_{(\downarrow 14.3)}$          & $53.4_{(\downarrow 15.6)}$           \\
			V                       & $50.4_{(\downarrow 23.4)}$        & $50.5_{(\downarrow 23.4)}$                & $41.4_{(\downarrow 26.7)}$          & $42.3_{(\downarrow 26.7)}$
			\\
			T+A                       & $71.6_{(\downarrow 2.2)}$        & $71.0_{(\downarrow 2.9)}$                & $66.3_{(\downarrow 1.8)}$          & $65.9_{(\downarrow 3.1)}$           \\
			T+V                       & $69.5_{(\downarrow 4.3)}$        & $68.7_{(\downarrow 5.2)}$                & $64.2_{(\downarrow 3.9)}$          & $64.1_{(\downarrow 4.9)}$           \\
			V+A                       & $63.7_{(\downarrow 10.1)}$        & $63.0_{(\downarrow 10.9)}$                &
			$54.6_{(\downarrow 13.5)}$          & $53.4 _{(\downarrow 15.6)}$          \\
			\hline
	\end{tabular}}
\end{table}

\textbf{Ablation studies for multimodal features.} We conduct ablation experiments on multimodal features to compare the performance of single-modal, bi-modal, and tri-modal experimental results to explore the importance of each modality. The experimental results are listed in Table \ref{tab:multimodal}. We choose W-Acc and W-F1 as evaluation metrics. In single-modal experiments, text modality features achieved the best performance, which shows that text features play a decisive role in MERC. Video features have the worst emotion recognition effect. We speculate that video features have more noise signals, making it difficult for the model to learn effective emotional feature representation. In bi-modal experiments, all bi-modal emotion recognition effects are better than their single-modal emotion recognition effects. The tri-modal emotion recognition effect is the best among all experiments. The performance improvement may be attributed to the effective fusion of multimodal, complementary semantic information, which can improve the feature representation ability of emotions. Therefore, GS-MCC can effectively utilize the consistent and complementary semantic information in multimodal conversations to improve the emotion recognition effect.

\section{Conclusions}
In this paper, we rethink the problem of multimodal emotion recognition in conversation from the perspective of the graph spectrum, taking into account some shortcomings of existing work and innovations. Specifically, we propose a Graph-Spectrum-based Multimodal Consistency and Complementary feature collaboration framework GS-MCC. First, we combine sliding windows to build a multimodal interaction graph to model the conversational relationship between utterances and speakers. Secondly, we design efficient Fourier graph operators to capture long-distance utterances' consistency and complementary semantic dependencies. Finally, we adopt contrastive learning and construct self-supervised signals with all negative samples to promote the collaboration of the two semantic information. Extensive experiments on two widely used benchmark datasets, IEMOCAP and MELD, demonstrate the effectiveness and efficiency of our proposed method.
	
\bibliographystyle{ACM-Reference-Format}
\bibliography{reference.bib}
	
\end{document}